# TRANSLATING MULTISPECTRAL IMAGERY TO NIGHTTIME IMAGERY VIA CONDITIONAL GENERATIVE ADVERSARIAL NETWORKS

*Xiao Huang[1]\*, Dong Xu[2], Zhenlong Li[1], Cuizhen Wang[1]*
[1]Department of Geography, University of South Carolina, Columbia, SC, U.S.A
[2]School of Geography Science, East China Normal University, Shanghai, China
\* Corresponding author: Xiao Huang (xh1@email.sc.edu)

## ABSTRACT

Nighttime satellite imagery has been applied in a wide range of fields. However, our limited understanding of how observed light intensity is formed and whether it can be simulated greatly hinders its further application. This study explores the potential of conditional Generative Adversarial Networks (cGAN) in translating multispectral imagery to nighttime imagery. A popular cGAN framework, pix2pix, was adopted and modified to facilitate this translation using gridded training image pairs derived from Landsat 8 and Visible Infrared Imaging Radiometer Suite (VIIRS). The results of this study prove the possibility of multispectral-to-nighttime translation and further indicate that, with the additional social media data, the generated nighttime imagery can be very similar to the ground-truth imagery. This study fills the gap in understanding the composition of satellite observed nighttime light and provides new paradigms to solve the emerging problems in nighttime remote sensing fields, including nighttime series construction, light desaturation, and multi-sensor calibration.

## 1. INTRODUCTION

The nocturnal lighting measured from space, serving as one of the hallmarks of human footprints on the Earth's surface, provides a unique perspective for revealing environmental and socio-economic dynamics [1]. Contrasting to daytime remote sensing, which largely depends on the physical characteristics of land cover, nighttime remote sensing is believed to render a new perspective on human activities, especially those activities associated with artificial light at night [2]. Over the past decades, the continuously updated nighttime sensors and improved nighttime products led to rich applications in a variety of fields [3,4].

Despite the popularity of nighttime products, however, limitations widely exist. Current nighttime remote sensing still lacks a long-time series of nighttime light observations, due to the inconsistency among/within sensors. The failure of designing efficient inter-calibration methods is largely due to our limited understanding of how nighttime lights are essentially formed, both physically and socioeconomically. In addition, the pixel-level relationship between nighttime lights and land features/human activities has not been fully explored [1]. How nighttime signals respond to different physical reflectance and socioeconomic activities is still unknown. The aforementioned knowledge gaps greatly hinder the further application of nighttime light data.

Fortunately, the latest development of Generative Adversarial Network (GAN) [5] provides a new opportunity for us to understand the formation of nighttime lights by allowing deep representations to be learned. As an emerging technique, GAN implicitly models high-dimensional distributions of data, characterized by two competing submodels: the generator and the discriminator. GAN can be extended to a conditional fashion, commonly known as conditional GANs (cGAN), in which the generator and the discriminator are conditioned on external information [6]. Such external information facilitates the deliberate or targeted generation of images. Proposed in 2014, cGAN has been applied in many image-to-image translation problems and achieved great performance. For example, pix2pix [7] proves the efficiency of cGAN by successfully translating a variety of image pairs, including map/aerial pair, edges/photo pair, and day/night pair. The great performance of cGAN in various image-to-image translation tasks raised a question: given sufficient training samples, can multispectral imagery be translated to nighttime imagery via cGAN?

To answer this question, this study explores the potential of cGAN in translating traditional multispectral imagery to nighttime imagery. Three scenarios were designed: 1) translating visible signals, i.e., RGB, to nighttime signals; 2) translating visible and infrared signals to nighttime signals; 3) translating visible and infrared signals, with the aid of social media distribution, to nighttime signals. In this study, a popular image-to-image translation framework, pix2pix [7], is adopted and modified to test the aforementioned three scenarios. The generated nighttime imagery and the ground-truth imagery are statistically compared via three evaluating metrics. This study investigates the possibility to translate multispectral imagery to nighttime imagery, largely benefiting a wide range of nighttime remote sensing studies by providing new paradigms for nighttime series construction, light desaturation, and multi-sensor calibration.

## 2. DATASETS

### 2.1 Multispectral imagery

The multispectral imagery used in this study is derived from Landsat 8 surface reflectance Tier 1 product, which has been atmospherically corrected and orthorectified. Queried via Google Earth Engine (GEE), the dataset covers the entire rectangular boundary of Conterminous U.S. (CONUS) with a temporal coverage from Jan 1st, 2016 to Dec 31st, 2016. A standard Landsat 8 cloud mask was implemented in GEE, and the median value was further selected for each pixel with overlapping values. The derived dataset contains four bands: Band 4 (red), Band3 (green), Band 2 (blue), and Band 5 (near-infrared). All the bands were resampled to a resolution of 100m.

### 2.2 Nighttime imagery

The nighttime imagery used in this study is derived from the Visible Infrared Imaging Radiometer Suite (VIIRS) Day-Night Band (DNB) carried by the Suomi National Polar-orbiting Partnership (NPP) Satellite, referred to as NPP-VIIRS. Launched in 2011, the NPP-VIIRS started to provide global observations (latitude ranges from 75N to 65S) of light intensity from 2012, with a nadir spatial resolution of 742m. To match the temporal coverage of Landsat 8 multispectral imagery, the version 1 annual stable light composite in the year 2016 was downloaded (https://ngdc.noaa.gov/eog/viirs/download_dnb_composites.html). The dataset has excluded any data impacted by stray light, lightning, lunar illumination, and cloud cover. Additional processing was done to screen out ephemeral lights and background (non-lights) [8]. To remove unusual intensity spikes in the dataset, we further preprocessed the NPP-VIIRS nighttime imagery by setting a maximum radiance of 300 $nW\ cm^{-2} sr^{-1}$ (values cap at 300 $nW\ cm^{-2} sr^{-1}$ if they are above 300 $nW\ cm^{-2} sr^{-1}$), a value found in the center of New York City and believed to be the highest urban light intensity.

### 2.3 Social media distribution

Twitter was selected as our social media platform, given its massive user base. Over 500 million geotagged tweets from the entire world from July 1st, 2016 to December 31st, 2016, were collected using Twitter Stream Application Programming Interface (API). These tweets were managed in a Hadoop (https:// hadoop.apache.org) environment and were then queried with Apache Impala (https://impala.incubator.apache.org). We further confined the distribution of tweets within the rectangular boundary of the CONUS and summarized the number of tweets in each 0.0042° by 0.0042° cell (approximately 450m by 450m at the equator). Given the great heterogeneity in Twitter distribution, log-transformation was applied for scaling purposes, followed by a low pass filter with a 3 by 3 kernel.

## 3. METHODOLOGY

### 3.1. Selection of training and validation grids

The rectangular boundary of CONUS is divided into a total of 40×96 grids, with each grid respectively spanning 0.625 in latitude and longitude. Due to the heterogeneous urban distribution in the CONUS, the majority of the grids are with limited light intensity, causing great difficulty for the multispectral-to-nighttime translation process. To select suitable training and validation grids, a threshold ($t$) of 8000 $nW\ cm^{-2} sr^{-1}$ was set. Grid with a total nighttime radiance above $t$ was determined as suitable, while grid with a total nighttime radiance below $t$ was excluded from the training process. A total of 1,000 grids were determined as suitable after the selection process and they were further randomly separated into a training dataset (800 grids) and a validation dataset (200 grids). The distribution of training grids and validation grids is presented in Fig. 1.

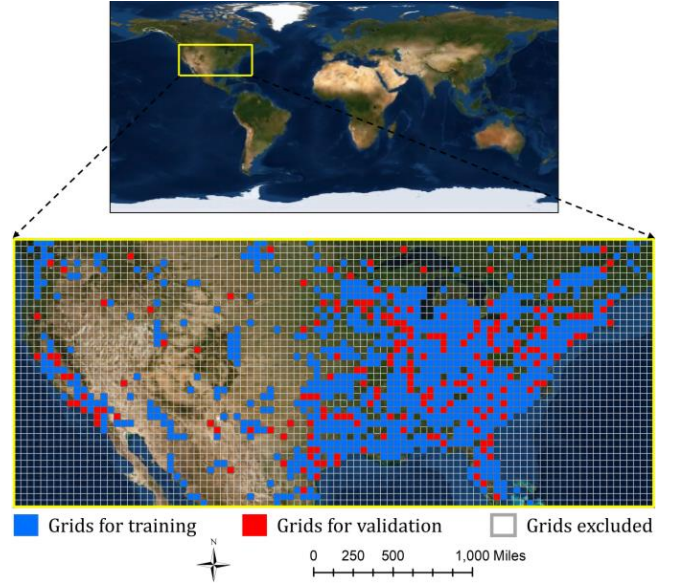

**Fig. 1.** Distribution of training and validation grids.

### 3.2. Multispectral-to-nighttime translation architecture

Given data $x$, latent vector $z$ and parameter $\theta$, GANs consist of two modules, one is the generator $G_{\theta_g}$ and the other is the discriminator $D_{\theta_d}$. The generator $G_{\theta_g}$ learns the data distribution from the latent vector $z$ and aims to produce realistic data, while the discriminator $G_{\theta_d}$ tries to distinguish whether the input is from the generator or the training data. Both modules compete to optimize their own objectives through a minimax two-player game with value function $\mathcal{L}_{cGAN}(\theta_g, \theta_d)$ [5]:

$$\min_{\theta_g} \max_{\theta_d} \mathcal{L}_{GAN}(\theta_g, \theta_d) = \min_{\theta_g} \max_{\theta_d} [\mathbb{E}_{x \sim p_{data}(x)} \log_{\theta_d}(x) + \mathbb{E}_{z \sim p(z)} \log(1 - D_{\theta_d}(G_{\theta_g}(z)))] \quad (1)$$

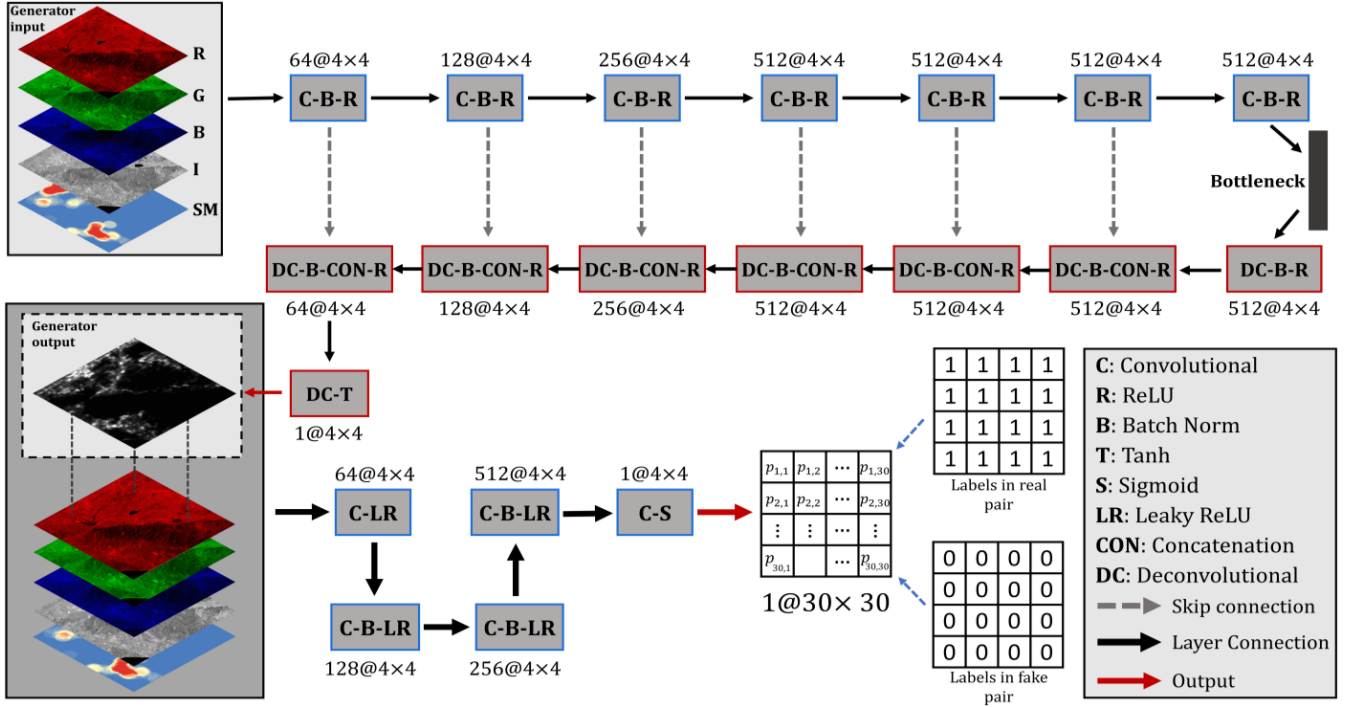

**Fig. 2.** The architecture of multispectral-to-nighttime translation in detail. R: Red, G: Green; B: Blue; I: Near-infrared; SM: Social Media

where generator $G_{\theta_d}$ tries to minimize $\mathbb{E}_{z \sim p(z)} \log(1 - D_{\theta_d}(G_{\theta_g}(z)))$, while the discriminator $D_{\theta_d}$ tries to maximize $\mathbb{E}_{x \sim p_{data}(x)} \log_{\theta_d}(x) + \mathbb{E}_{z \sim p(z)} \log(1 - D_{\theta_d}(G_{\theta_g}(z)))$.

Given the supervised nature of our proposed problem (i.e., translating multispectral imagery to nighttime imagery), additional condition ($\alpha$) needs to be observed by the discriminator $D_{\theta_d}$. The value function $\mathcal{L}_{GAN}(\theta_g, \theta_d)$, conditioned on $\alpha$, can be represented as:

$$\min_{\theta_g} \max_{\theta_d} \mathcal{L}_{GAN}(\theta_g, \theta_d) = \min_{\theta_g} \max_{\theta_d} [\mathbb{E}_{\alpha, x \sim p_{data}(\alpha,x)} \log_{\theta_d}(\alpha, x) + \mathbb{E}_{\alpha \sim p_{data}(\alpha), z \sim p(z)} \log(1 - D_{\theta_d}(\alpha, G_{\theta_g}(z)))] \quad (2)$$

where the generator $G_{\theta_d}$ tries to minimize $\mathbb{E}_{\alpha \sim p_{data}(\alpha), z \sim p(z)} \log(1 - D_{\theta_d}(\alpha, G_{\theta_g}(z)))$, while the discriminator $D_{\theta_d}$ tries to maximize $\mathbb{E}_{\alpha, x \sim p_{data}(\alpha,x)} \log_{\theta_d}(\alpha, x) + \mathbb{E}_{\alpha \sim p_{data}(\alpha), z \sim p(z)} \log(1 - D_{\theta_d}(\alpha, G_{\theta_g}(z)))$. To force the generated nighttime imagery for given multispectral imagery to remain as similar as possible to the corresponding input pairs (the ground truth), an L1 loss is introduced to the generator $G_{\theta_d}$:

$$\mathcal{L}_{L1}(\theta_g) = \mathbb{E}_{\alpha, x \sim p_{data}(\alpha,x), z \sim p(z)} \|x - G_{\theta_g}(\alpha, z)\| \quad (3)$$

Together with the conditional GAN loss, the final loss ($G^*$) can be expressed as:

$$G^* = \min_{\theta_g} \max_{\theta_d} [\mathcal{L}_{GAN}(\theta_g, \theta_d) + \lambda \mathcal{L}_{L1}(\theta_g)] \quad (4)$$

where $\lambda$ denotes a weighting term for L1 loss. It was set to 100, according to [7].

Fig. 2 presents the architecture of multispectral-to-nighttime translation in detail. The generator features a "U-Net" design [9] with skip connections, concatenating all channels at layer $i$ with those at layer $n - i$, where $n$ denotes the total number of layers. The discriminator adopts the concept of PatchGAN, a convolutional neural network that classifies each 70×70 patch in the original image as real or fake. Each block of the discriminator follows the "Convolution-BatchNorm-ReLU" design. The only exception is the first layer of the PatchGAN, where batch normalization is not utilized.

### 3.3. Evaluating metrics

Three evaluating metrics were used to compare the generated nighttime imagery with ground-truthing nighttime imagery. Those metrics include Euclidean Distance ($D_{eu}$), Manhattan Distance ($D_{ma}$) and Normalized Cross-correlation ($R_{ncc}$):

$$D_{eu} = \sqrt{\sum_{(i,j) \in R^{256 \times 256}} (G_{i,j} - O_{i,j})^2} \quad (5)$$

$$D_{ma} = \sum_{(i,j) \in R^{256 \times 256}} |G_{i,j} - O_{i,j}| \quad (6)$$

$$R_{ncc} = \frac{\sum_{(i,j) \in R^{256 \times 256}} (G_{i,j} - \bar{G})(O_{i,j} - \bar{O})}{\sqrt{\sum_{(i,j) \in R^{256 \times 256}} (G_{i,j} - \bar{G})^2 \sum_{(i,j) \in R^{256 \times 256}} (O_{i,j} - \bar{O})^2}} \quad (7)$$

where $G_{i,j}$ and $O_{i,j}$ represent the standardized pixel value (in a range of [-1,1]) in row $i$ and column $j$ of generated imagery and ground-truth imagery, respectively. Both generated and ground-truth imagery are compared within the domain $R$, a two-dimensional space of 256×256.

## 4. RESULTS

Fig. 3 presents the multispectral-to-nighttime translation results for four example sites using three different scenarios. Visually compared with the ground-truth nighttime imagery, the translations using only RGB reflectance achieved the worst performance as the generated nighttime imagery failed to well agree with the ground-truth. With the additional input of the infrared signal, the translation performance was significantly boosted, evidenced by the detailed identification of urban fabrics and the removal of noises in densely vegetated areas. This improvement is presumably due to the large discrepancy in infrared reflectance between man-made objects and vegetation, allowing proper and efficient features to be learned by the proposed model to better distinguish human settlements. In addition, the involvement of social media distribution further improved the translation performance as the generated nighttime imagery only shows small discrepancies compared with the ground-truth. Social media aids the translation process by providing a unique social perspective in addition to the already captured physical characteristics, leading to better recognition of urban centers and urban functional cores in the translated imagery.

Table. 1 presents the translation performance of the three evaluating metrics for the 200 validation grids. $D_{eu}$ and $D_{ma}$ quantitively measure the distance between two images (the smaller, the better), while $R_{ncc}$ measures the degree of similarity between two images (1 as identical). All the three evaluating metrics confirm that the scenario using physical reflectance of RGB and infrared, together with the involvement of social media distribution, i.e., the RGBISM, achieved the best performance.

## 5. CONCLUSIONS

This article explores the possibility of translating multispectral remote sensing imagery to nighttime remote sensing imagery via conditional generative adversarial networks. The results prove the feasibility of this multispectral-to-nighttime translation task and further indicate that, with the integration of infrared signal and additional social media data, the generated nighttime imagery can be very close to the ground-truth. The results of this study are expected to benefit a wide range of nighttime remote sensing studies by providing new paradigms to solve the emerging problems, including nighttime series construction, light desaturation, and multi-sensor calibration.

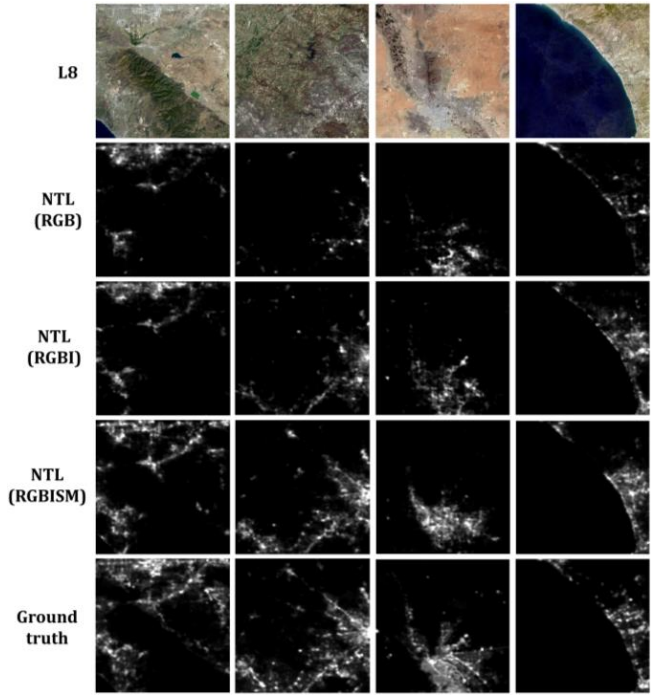

**Fig. 3.** Multispectral-to-nighttime imagery translation results in sample grids

**Table 1.** Performance of three scenarios in the multispectral-to-nighttime translation task

| Evaluating metrics | Scenarios | | |
|---|---|---|---|
| | RGB | RGBI | RGBISM |
| $D_{eu}$ | 31.431 | 24.872 | **19.214** |
| $D_{ma}$ | 41.947 | 31.826 | **23.234** |
| $R_{ncc}$ | 0.492 | 0.647 | **0.821** |

*Note.* The best results in the metrics are highlighted in bold.